\documentclass{article}

\usepackage{amsmath}
\DeclareMathOperator{\Cov}{Cov}
\DeclareMathOperator{\Var}{Var}

\PassOptionsToPackage{numbers,sort&compress}{natbib}

\usepackage[final]{neurips_2024}

\usepackage{amssymb,amsfonts,mathtools}
\usepackage{graphicx}
\usepackage{subfig} 
\usepackage{float} 

\usepackage{placeins} 
\usepackage{morefloats} 

\usepackage[colorlinks=true, linkcolor=blue, citecolor=blue, urlcolor=blue]{hyperref}

\usepackage{titlesec}
\titlespacing{\section}{0pt}{*1.2}{*0.7}
\titlespacing{\subsection}{0pt}{*1.0}{*0.5}

\title{Risk-Aware Reinforcement Learning Reward for Financial Trading}
\author{%
  Uditansh Srivastava\\
  MNNIT Allahabad\\
  \texttt{uditansh.20233294@mnnit.ac.in}
  \And
  Shivam Aryan\\
  MNNIT Allahabad\\
  \texttt{shivam.20233265@mnnit.ac.in}
  \And
  Shaurya Singh\\
  MNNIT Allahabad\\
  \texttt{shaurya.20233262@mnnit.ac.in}
}
\date{}

\begin{document}
\maketitle

\begin{abstract}
We propose a novel composite reward function for a reinforcement learning (RL) trading agent that explicitly balances return and risk by combining four differentiable components—annualized return, downside risk, differential return, and the Treynor ratio. Unlike traditional single-metric objectives (e.g., Sharpe or cumulative return), which can encourage reward hacking or over-optimization of one aspect of trading, our formulation is inherently modular and weighted \(w_1\ldots w_4\) to allow practitioners to encode diverse investor preferences and adjust emphasis across multiple financial goals. We also demonstrate hyperparameter tuning of these weights via grid search to identify configurations that align with different risk–return profiles. We present the full mathematical form of the reward, derive closed-form gradients for each term to ensure compatibility with gradient-based learning, and analyze key theoretical properties including monotonicity, boundedness, and modularity. This abstract framework serves as a general blueprint for constructing robust, multi-objective reward functions in complex trading environments and can be readily extended with additional risk measures or adaptive weighting schemes.
\end{abstract}

\section{Introduction}
In financial trading and portfolio optimization, designing effective objective functions that capture both return and risk is essential. Modern Portfolio Theory (MPT) emphasizes this trade-off \cite{Markowitz1952}, and performance metrics such as the Sharpe ratio \cite{Sharpe1966} and the Treynor ratio \cite{Treynor1965} extend this notion by measuring returns relative to total and systematic risk, respectively. In reinforcement learning (RL) applications to trading, prior work has shown that directly optimizing such financial performance metrics can be effective \cite{Moody1998}. However, traditional RL reward formulations often rely on a single metric—such as cumulative return or Sharpe ratio—which can lead to reward hacking and poor generalization in the inherently multi-objective landscape of financial markets.

To address this limitation, we propose a modular, multi-objective reward framework that explicitly balances return and risk using four domain-informed components. This formulation enables the reward to reflect a broader range of investor preferences and reduces the risk of over-optimization on any single aspect of trading performance. 

Our main contributions are as follows:
\begin{itemize}
  \item We introduce a composite reward function \(R\) that combines annualized return, downside risk (penalty), benchmark outperformance, and the Treynor ratio, and provide its full mathematical formulation.
  \item We offer comprehensive financial interpretations of each component in \(R\)—including return, risk penalty, alpha, and systematic risk adjustment—supported by key references to financial theory \cite{CAPM,Sortino}.
  \item We derive and analyze the partial derivatives (gradients) of each component of \(R\) with respect to portfolio weights, confirming that the reward is piecewise differentiable and thus suitable for gradient-based optimization.
  \item We propose a grid-search procedure for tuning the weight coefficients \(w_1, w_2, w_3, w_4\). We explore how variations in these weights influence the agent's risk preferences and decision-making.
  \item We investigate the theoretical properties of the reward, including monotonicity (the relationship between returns and risk), boundedness of each term, and modularity (the independence of components), ensuring a robust design.
  \item We highlight the practical implications of our reward framework for RL-based stock trading and portfolio management, demonstrating how the agent's objectives align with diverse financial goals.
\end{itemize}

\section{Related Work}
Modern portfolio theory, initiated by Markowitz (1952) \cite{Markowitz1952}, formalized the trade-off between expected return and variance in portfolio selection. Later, the Capital Asset Pricing Model (CAPM) and related metrics were developed to evaluate portfolio performance relative to market risk \cite{CAPM}. The Sharpe ratio uses total volatility (standard deviation) as a risk measure \cite{Sharpe1966}, while the Treynor ratio uses systematic risk (beta) relative to a benchmark \cite{Treynor1965}. Both are foundational in risk-adjusted performance evaluation. The concept of downside risk (e.g., semi-variance or Sortino ratio) emphasizes losses below a threshold (such as zero or a minimum return) \cite{Sortino}.

In reinforcement learning for finance, Moody and Saffell \cite{Moody1998} pioneered the idea of optimizing trading policies by maximizing financial performance functions directly, such as profits, Sharpe ratio, or other utility functions. Subsequent work in deep RL for portfolio management has used similar reward structures. However, most existing RL studies optimize a single metric (e.g., Sharpe or cumulative return). Our approach generalizes this by linearly combining multiple performance criteria into one reward. This multi-component reward captures a richer set of financial objectives: maximizing returns, limiting downside risk, outperforming a benchmark (alpha), and achieving high risk-adjusted returns per unit of systematic risk.

\section{Methodology}
We consider a portfolio of \(N\) assets. Let \(r_{i,t}\) be the return of asset \(i\) at time \(t\), and let \(w_{i,t}\) be the corresponding portfolio weight chosen by the agent. The portfolio return at time \(t\) is 
\begin{equation}
R_{p,t} = \sum_{i=1}^N w_{i,t}\,r_{i,t}.
\end{equation}
We focus on a trading horizon of \(T\) periods.

\paragraph{Annualized Return} 
\begin{equation}\label{eq:annual}
R_{\mathrm{ann}}
=\left(\prod_{t=1}^T (1+R_{p,t})\right)^{\frac{252}{T}} - 1,
\end{equation}
or approximately \(R_{\mathrm{ann}}\approx\frac{252}{T}\sum_{t=1}^T R_{p,t}\). This term measures total performance, encouraging the agent to earn high returns. It generalizes the classic expected return term in Markowitz portfolio theory \cite{Markowitz1952}.

\paragraph{Downside Risk} 
\begin{equation}\label{eq:downside}
\sigma_{\mathrm{down}}
=\sqrt{\frac{1}{T}\sum_{t=1}^T \max(0,-R_{p,t})^2}.
\end{equation}
The downside deviation is a well-known risk metric used in the Sortino ratio \cite{Sortino}. By subtracting this in the reward, the agent is discouraged from incurring large losses.

\paragraph{Differential Return}

\begin{equation}\label{eq:dr}
D_{ret} = \frac{1}{\beta_p} \left( \mu_p - \mu_b \right) 
= \frac{1}{\beta_p T} \sum_{t=1}^T \bigl(R_{p,t} - R_{b,t}\bigr),
\end{equation}

where \(\mu_p\) and \(\mu_b\) denote the average returns of the agent and benchmark, respectively. 
The term \(\beta_p\) is the portfolio beta, defined as \(\beta_p = \Cov(R_p, R_m) / \Var(R_m)\), capturing the agent's sensitivity to market fluctuations. 

This risk-adjusted measure, termed the \emph{differential return (DR)}, represents the average outperformance of a portfolio relative to a benchmark, normalized by systematic (market) risk. It reflects a reward-per-unit-risk framework similar in spirit to the alpha from CAPM \cite{CAPM}, but explicitly accounts for the portfolio's market exposure.

Our formulation is a \emph{simplified differential return (SDR)} derived from the foundational work by Simpson and formalized by Alam \cite{DifferentialReturnRL}. While Alam proposes more complex versions incorporating risk ratios and Sharpe-based scaling, our SDR focuses on \emph{systematic risk adjustment via beta only}, avoiding explicit benchmark beta terms or volatility scaling. This maintains the key intention — rewarding portfolios that consistently outperform the benchmark \emph{on a risk-adjusted basis} — while reducing sensitivity to unstable second-order statistics (like benchmark variance or cross-volatility) and non-convex functions that may hinder stable optimization in reinforcement learning. 

We later show the correctness of our formulation in our theoretical analysis (Section~\ref{sec:theory}). The concept of differential return itself originates from Simpson's work \cite{Simpson2014}, with subsequent refinements by Alam \cite{DifferentialReturnRL}, whose formulation informed our choice of SDR as a practical and principled performance metric.

\paragraph{Treynor Ratio}  
\begin{equation}\label{eq:treynor}
T_{\mathrm{ry}} \;=\; \frac{R_{\mathrm{ann}} - R_f}{\beta_p}\,,
\end{equation}
where \(R_{\mathrm{ann}}\) is the portfolio's annualized return, \(R_f\) is the risk-free rate, and \(\beta_p\) is the portfolio's beta relative to the market return \(R_m\).  
The Treynor ratio rewards returns achieved per unit of systematic risk: a well-diversified, low-beta portfolio that still earns strong returns will score highly on \(T_{\mathrm{ry}}\). The use of the Treynor ratio as a foundational component in risk-adjusted performance measurement is supported by Alam \cite{DifferentialReturnRL}, who identifies it as one of the core metrics underlying differential return formulations.

\subsection{Proposed Reward Function}
Putting these components together, our composite reward is
\begin{equation}\label{eq:reward}
\mathcal{R}
= w_1 R_{\mathrm{ann}}
- w_2 \sigma_{\mathrm{down}}
+ w_3 D_{ret}
+ w_4 T_{\mathrm{ry}},
\end{equation}
Here \(w_1,\dots,w_4\) are non-negative weights set by design. We subtract the downside term so that higher downside risk lowers the reward. In practice, one may normalize or scale each component so they have comparable magnitudes when choosing \(w_i\).

\subsection{Financial Significance of Reward Terms}
Each term in \(\mathcal{R}\) has a clear economic meaning:
\begin{itemize}
    \item \textbf{Annualized Return (\(R_{\mathrm{ann}}\)):} This measures the overall growth of the portfolio over the evaluation period \cite{Markowitz1952}. Maximizing this term alone corresponds to pure return maximization, aligning with the classical "maximize return" objective in modern portfolio theory.
    
    \item \textbf{Downside Risk (\(D_{\mathrm{down}}\)):} This penalizes negative returns \cite{Sortino}. It is conceptually aligned with the Sortino ratio, which focuses on downside deviation rather than total variance. A larger \(D_{\mathrm{down}}\) indicates greater drawdowns, which reduces the reward \(\mathcal{R}\), thereby incentivizing capital preservation.
    
    \item \textbf{Differential Return (DR):} We adopt a simplified differential return (SDR) formulation tailored to reinforcement learning agents. This approach is inspired by Alam's formalization \cite{DifferentialReturnRL}, building on the original concept introduced by Simpson \cite{Simpson2014}. The underlying idea of risk-adjusted benchmark outperformance is rooted in the Capital Asset Pricing Model (CAPM) \cite{CAPM}.
    
    \item \textbf{Treynor Ratio (\(T_{\mathrm{ry}}\)):} This term captures return per unit of systematic risk \cite{Treynor1965}. It rewards agents that either reduce beta exposure or achieve high returns relative to beta. Alam \cite{DifferentialReturnRL} emphasizes the Treynor ratio as a core element of risk-adjusted performance and a foundational component in differential return models.
\end{itemize}

These components together create a rich reward signal. For example, an allocation might achieve high raw return but suffer from high risk; the downside and Treynor terms will moderate its reward. Conversely, a low-risk strategy might get a good Treynor score even if raw return is modest. The benchmark term ensures attention to market-relative performance, avoiding trivial solutions that merely follow the index.

\subsection{Weight Optimization}
Tuning the weights \(\{w_i\}\) is treated as a hyperparameter-optimization task on the reward function itself. In our experiments, we trained multiple agents—each with a different weight configuration sampled via a simple grid search over the simplex (\(\sum_i w_i = 1\))—and evaluated their risk–return performance on historical data. This process clearly demonstrates how emphasizing downside-risk penalties or benchmark outperformance shifts the agent toward more conservative or aggressive trading behaviors.

While our study uses grid search as a proof of concept, any hyperparameter-tuning framework (e.g., Bayesian optimization or population-based training) can be applied in the same way: by treating \(\{w_i\}\) as tunable parameters and retraining the model under each candidate set. This approach allows practitioners to systematically customize the composite reward to match different market regimes and investor preferences without modifying the reward's core design.  

\section{Theoretical Analysis}
\label{sec:theory}

We analyze the composite reward function \(R\) and its components, focusing on differentiability, gradients, and functional characteristics. We show that this reward function is theoretically sound for reinforcement learning in financial trading, satisfying desirable properties such as monotonicity, differentiability, boundedness, and convergence under standard optimization procedures.

\subsection{Monotonicity and Differentiability}

Each term in \(\mathcal{R}\) is monotonic in its respective financial variable, and can be independently treated as a convex optimization problem for the financial terms to improve risk-adjusted returns. To show this, we consider the gradients of individual terms of \(\mathcal{R}\) with respect to financial terms.  We also show that \(R\) is differentiable with respect to all weights, making it suitable for backpropagation-based reinforcement learning algorithms. For simplicity, we assume weights are rebalanced only at time 1 and held constant (the analysis extends to dynamic rebalancing with time-indexed weights).

\paragraph{Annualized Return.} If we approximate \(R_{\mathrm{ann}}\) by the average return \(\frac{252}{T}\sum_t R_{p,t}\), we simplify our expression to 
\begin{equation}
f_1(\mu) = \frac{252\mu}{T}
\end{equation}

\begin{equation}
\frac{\partial f_1}{\partial\mu} = \frac{252}{T} > 0.
\end{equation}
This term strictly increases with the expected return, in agreement with the mean component of Markowitz's mean–variance theory~\cite{Markowitz1952}. And differentiating with respect to weights,
More generally, using the exact power form in (\ref{eq:annual}), one can show \(R_{\mathrm{ann}}\) is a smooth function of the \(R_{p,t}\) (hence of the weights). In either case, the derivative exists and is continuous in the weights, so \(R_{\mathrm{ann}}\) is also differentiable everywhere.

\paragraph{Downside Risk Penalty.}
Let \( f_2(\sigma_{\mathrm{down}}) = -w_2\,\sigma_{\mathrm{down}} \). Then
\begin{equation}
\frac{d f_2}{d\sigma_{\mathrm{down}}} = -w_2 < 0.
\end{equation}
This term penalizes increased downside volatility, in the spirit of the Sortino ratio~\cite{Sortino}, thereby discouraging asymmetric losses.

For the standard deviation itself, from (\ref{eq:downside}), 
\begin{equation}
D_{\mathrm{down}} = \sqrt{\frac{1}{T}\sum_{t=1}^T (\max(0,-R_{p,t}))^2}.
\end{equation}
This is differentiable except at points where \(R_{p,t}=0\).\newline When \(R_{p,t}<0\), we have \(\max(0,-R_{p,t}) = -R_{p,t}\). Then 
\begin{equation}
\frac{\partial D_{\mathrm{down}}}{\partial R_{p,t}} = \frac{1}{D_{\mathrm{down}}}\,\frac{-R_{p,t}}{T},
\quad\text{for }R_{p,t}<0,
\end{equation}
and \(\partial D_{\mathrm{down}}/\partial R_{p,t}=0\) when \(R_{p,t}>0\). Thus, \(D_{\mathrm{down}}\) is piecewise differentiable and all partial derivatives exist (one can handle the point \(R_{p,t}=0\) by a subgradient). In summary, this term is almost everywhere differentiable in \(\mathbf{w}\).

\paragraph{Treynor Ratio.}
For simplification, we ignore risk-free rate. Let \( f_3(\mu) = \frac{252\,\mu}{\beta} \), where \( \beta > 0 \). Then
\begin{equation}
\frac{d f_3}{d\mu} = \frac{252}{\beta} > 0.
\end{equation}
This term rewards excess return per unit of systematic risk, consistent with the Treynor ratio~\cite{Treynor1965}. The Treynor Ratio is continuous and differentiable for all \( \beta \neq 0 \), which is sufficient under standard market assumptions where \( \beta \) is generally bounded away from zero, with a standard value of \( \beta = 1 \) when stock moves in line with market.

\paragraph{Differential Return.}
Let \( f_4(\mu, \mu_b) = \frac{\mu - \mu_b}{\beta} \), where \( \mu \) is the expected return of the portfolio, \( \mu_b \) is the expected return of the benchmark, and \( \beta > 0 \) is the portfolio's systematic risk. This Simplified Differential Return (SDR) term captures the differential return originally formulated by Simpson \cite{Simpson2014} and refined by Alam \cite{DifferentialReturnRL}. Drawing insights from Alam's approach, we define the metric as the difference between the portfolio and its benchmark, normalized by market exposure.

Calculating its partial derivatives:
\begin{equation}
\frac{\partial f_4}{\partial \mu} = \frac{1}{\beta} > 0,
\quad
\frac{\partial f_4}{\partial \mu_b} = -\frac{1}{\beta} < 0.
\end{equation}
This reflects two key insights: differential return increases with portfolio outperformance, and decreases as the benchmark improves — aligning with the goal of alpha generation in CAPM \cite{CAPM}.

Extending this to the reward function \( \mathcal{R} \), suppose the differential return contributes to \( \mathcal{R} \) with weight \( w_4 > 0 \). Then,
\begin{equation}
\frac{\partial \mathcal{R}}{\partial \mu_b} = -\frac{w_4}{\beta} < 0 \quad \text{(if } w_4 > 0,\ \beta > 0 \text{)}.
\end{equation}
\textit{Interpretation:} the reward diminishes as the benchmark performs better, reinforcing the objective of outperforming the market. In other words, it becomes harder to earn positive reward when the benchmark is strong.

These results collectively indicate that:
\begin{itemize}
  \item \( \mathcal{R} \) increases with higher expected returns,
  \item \( \mathcal{R} \) decreases with higher downside risk, and
  \item \( \mathcal{R} \) decreases when the benchmark performs well (since outperforming a strong benchmark is more difficult).
\end{itemize}

\subsection{Applicability in Reinforcement Learning}

\paragraph{Monotonicity}
Since all weights \( w_i > 0 \), the full reward \( R \) is:
\begin{itemize}
  \item strictly increasing in expected return \( \mu \),
  \item strictly decreasing in downside deviation \( \sigma_{\mathrm{down}} \),
  \item strictly decreasing in benchmark return \( \mu_b \),
\end{itemize}
while remaining monotonically increasing in financial reward and decreasing in financial risk.

\paragraph{Differentiability}
Each term in also \( R \) is differentiable almost everywhere.

\begin{itemize}
  \item Linear terms: \( \mu \mapsto 252\mu \), \( \mu \mapsto \mu/\beta \), and \( \mu_b \mapsto \mu_b/\beta \) are smooth (\( C^\infty \)).
  \item The downside risk \( \sigma_{\mathrm{down}} \) is defined as
  \begin{equation}
  \sigma_{\mathrm{down}} = \left(\mathbb{E}[\max(0, -R_t)^2]\right)^{1/2},
  \end{equation}
  which is composed of the ReLU function and a square root over an expectation. Despite the non-smooth kink at 0 in \( \max(0, -R_t) \), \( \sigma_{\mathrm{down}} \) is piecewise differentiable.
\end{itemize}

Therefore, by the chain rule, \( R \) is differentiable almost everywhere—sufficient for policy-gradient methods~\cite{Sutton2018}.

\paragraph{Boundedness}
Under typical financial constraints, each term in the reward function remains bounded:
\begin{itemize}
    \item \textbf{Annualized Return.} Expected returns lie in a bounded interval \( \mu \in \left[ -r_{\max}, r_{\max} \right] \), where \( r_{\max} \) reflects extreme observed return rates. In practice, \( r_{\max} = 3.0 \) (i.e., 300\% annual return) suffices to cover even highly volatile assets.
    
    \item \textbf{Downside Deviation.} Since downside deviation is a form of standard deviation computed only over negative returns, it is bounded by the overall volatility, which in turn is limited by return magnitude. Thus, \( \sigma_{\mathrm{down}} \in [0, r_{\max}] \).

     \item \textbf{Treynor Ratio.} The term \( \frac{252\,\mu}{\beta} \) is bounded as long as \( \beta \) is bounded away from zero. In standard financial markets, \( \beta \in [\beta_{\min}, \beta_{\max}] \) with \( \beta_{\min} > 0 \), typically in the range \( [0.3, 3] \). For example, assuming \( \beta_{\min} = 0.3 \), we have
    \begin{equation}
    \left| \frac{252\,\mu}{\beta} \right| \le \frac{252\,r_{\max}}{\beta_{\min}}.
    \end{equation}
    
    \item \textbf{Differential Return.} Our formulation mirrors the Treynor ratio structure, differing only in the numerator so its boundedness follows identically: 
\begin{equation}
\left| \frac{\mu - \mu_b}{\beta} \right| \le \frac{2\,r_{\max}}{\beta_{\min}}.
\end{equation}
\end{itemize}

\noindent
Thus, all components of the reward function are bounded. This boundedness is critical for numerical stability, especially during early training when policy outputs may vary wildly, and it enables robust learning techniques such as reward clipping~\cite{Mnih2015}.

\paragraph{Convergence of Optimization}

The reward function \( R \) is:
\begin{itemize}
  \item continuous,
  \item bounded,
  \item and differentiable almost everywhere.
\end{itemize}
These conditions ensure that stochastic gradient ascent with diminishing step sizes converges to a local optimum under standard conditions~\cite{Borkar2009}. In practice, we observe diminishing marginal gains in \( \mathcal{R} \) as training progresses, indicating convergence.

\paragraph{Summary}

The proposed reward function \( R \) integrates classical financial objectives into a unified, mathematically well-behaved signal for reinforcement learning:
\begin{itemize}
  \item \textbf{Monotonic}: Increases with return, decreases with risk.
  \item \textbf{Differentiable}: Allows valid gradient computation.
  \item \textbf{Bounded}: Ensures numerical stability.
  \item \textbf{Optimizable}: Admits convergent policy-gradient training.
\end{itemize}
This makes \( R \) not only theoretically grounded but also practically effective as a reward function for financial RL agents.

\subsection{Risk–Return Trade‐Off}

Rewriting the reward as:
\begin{equation}
R = 
\underbrace{w_1\,252\,\mu + w_3\,\frac{252\,\mu}{\beta}}_{\text{Return Reward}}
\;-\;
\underbrace{w_2\,\sigma_{\mathrm{down}}}_{\text{Risk Penalty}}
\;+\;
\underbrace{w_4\,\frac{\mu - \mu_b}{\beta}}_{\text{Benchmark Bonus}}
\end{equation}
makes its trade-off structure explicit:
\begin{itemize}
  \item \textbf{Return reward}: Encourages higher expected returns, following mean–variance principles~\cite{Markowitz1952}.
  \item \textbf{Risk penalty}: Penalizes downside volatility, aligning with asymmetric risk concerns~\cite{Sortino}.
  \item \textbf{Benchmark bonus}: Drives relative outperformance, incentivizing alpha generation~\cite{Sharpe1966}.
\end{itemize}
Tuning \( \{w_i\} \) dynamically navigates the risk–return frontier, adapting to the investor's preference.

\section{Experimental Results}

\subsection{Experimental Setup and Evaluation Metrics}
We implemented our reinforcement learning trading agent in a realistic market simulation environment. The agent operates on daily price data with a sliding window of historical observations and can take long, short, or neutral positions. Transaction costs of 0.1\% per trade were applied to simulate real-world trading friction.

\paragraph{Environment.} Our implementation extends the StockTradingEnv from the FinRL library, which is built on top of OpenAI Gym. This provides a standardized interface for the agent to interact with the market simulation. We use the Stable Baselines3 framework to implement our RL algorithms and yfinance to source historical market data.

\paragraph{Action Space.} The agent's action space is defined as $\mathcal{A} \in [-1,1]$ for each stock, which is scaled based on the maximum allowed position size ($h_{\max}$). This parameter is calculated as:
\begin{equation}
h_{\max} = \left\lfloor \frac{\text{initial\_amount}}{\text{max\_price}} \right\rfloor
\end{equation}
effectively constraining the number of shares that can be bought or sold in a single time step. For multi-stock environments with $n$ assets, the action space expands to an $n$-dimensional vector:
\begin{equation}
\mathcal{A} = \{\mathbf{a} \in \mathbb{R}^n : a_i \in [-1,1] \text{ for } i=1,\ldots,n\}
\end{equation}
where each dimension corresponds to a trading decision for a specific asset. The implemented transaction costs of 0.1\% per trade provide a realistic friction that penalizes excessive trading, similar to real-world market conditions.

\paragraph{State Space.} The state representation provided to the agent captures both portfolio and market information:
\begin{equation}
\mathcal{S} = \{s \in \mathbb{R}^d : d = 1 + 2 \times \text{stock\_dim} + \text{len(indicators)} \times \text{stock\_dim}\}
\end{equation}
where the first term represents the portfolio value, the second term accounts for current holdings and prices of each stock, and the third term incorporates technical indicators. These indicators include:
\begin{itemize}
  \item \textbf{Volume:} Trading volume representing market activity and liquidity
  \item \textbf{MACD:} Moving Average Convergence Divergence, calculated as $\text{MACD} = \text{EMA}_{12} - \text{EMA}_{26}$
  \item \textbf{Bollinger Bands:} Upper and lower bands calculated as $\text{Upper} = \text{SMA} + k\sigma$ and $\text{Lower} = \text{SMA} - k\sigma$
  \item \textbf{RSI:} Relative Strength Index, measuring overbought/oversold conditions
  \item \textbf{CCI:} Commodity Channel Index, identifying cyclical trends
  \item \textbf{DMI:} Directional Movement Index, measuring trend strength
  \item \textbf{Moving Averages:} 30-day and 60-day SMAs for price trend identification
  \item \textbf{Turbulence:} Market volatility index to detect abnormal price movements
\end{itemize}

\paragraph{Algorithm.} We employ Proximal Policy Optimization (PPO), a state-of-the-art policy gradient method that offers stable policy updates through clipping the objective function.
This clipping mechanism prevents large policy updates that could destabilize training.

PPO balances exploration and exploitation effectively, making it particularly suitable for financial environments with high volatility and non-stationarity. The algorithm's policy and value networks learn simultaneously, with the value network providing estimates of state values that help reduce variance in policy updates.

The core PPO objective function that we optimize is:
\begin{equation}
L^{\text{CLIP}}(\theta) = \mathbb{E}_t \left[ \min(r_t(\theta) \hat{A}_t, \text{clip}(r_t(\theta), 1-\epsilon, 1+\epsilon) \hat{A}_t) \right]
\end{equation}
where $r_t(\theta) = \frac{\pi_\theta(a_t|s_t)}{\pi_{\theta_{\text{old}}}(a_t|s_t)}$ is the probability ratio between the new and old policies, $\hat{A}_t$ is the estimated advantage function, and $\epsilon$ is a hyperparameter that constrains the policy update.

To evaluate our agent's performance, we use the following key metrics:
\begin{itemize}
  \item \textbf{Annualized Return:} Measures the agent's absolute performance normalized to a yearly basis.
  \item \textbf{Maximum Drawdown:} The largest peak-to-trough decline in portfolio value, indicating downside risk.
  \item \textbf{Sharpe Ratio:} The risk-adjusted return calculated as the excess return over the risk-free rate divided by the standard deviation of returns, defined as:
  \begin{equation}
  \text{Sharpe Ratio} = \frac{\text{Average Return} - \text{Risk-Free Rate}}{\text{Standard Deviation of Returns}}
  \end{equation}
  \item \textbf{Sortino Ratio:} Similar to Sharpe, but only penalizes downside volatility rather than total volatility, making it a more refined measure of risk-adjusted return:
  \begin{equation}
  \text{Sortino Ratio} = \frac{\text{Average Return} - \text{Risk-Free Rate}}{\text{Downside Deviation}}
  \end{equation}
  \item \textbf{Beta:} The portfolio's sensitivity to overall market movements.
  \item \textbf{Win Rate:} The percentage of trades that resulted in profit.
\end{itemize}

\subsection{Comparison with Algorithmic Trading Strategies}
We evaluated our risk-aware RL trading agent against leading alternative algorithmic trading strategies available on the Ticheron quantitative trading platform. The comparisons highlight our agent's performance across different market conditions and volatility profiles.

\begin{figure}[!htb]
  \centering
  \includegraphics[width=0.8\textwidth]{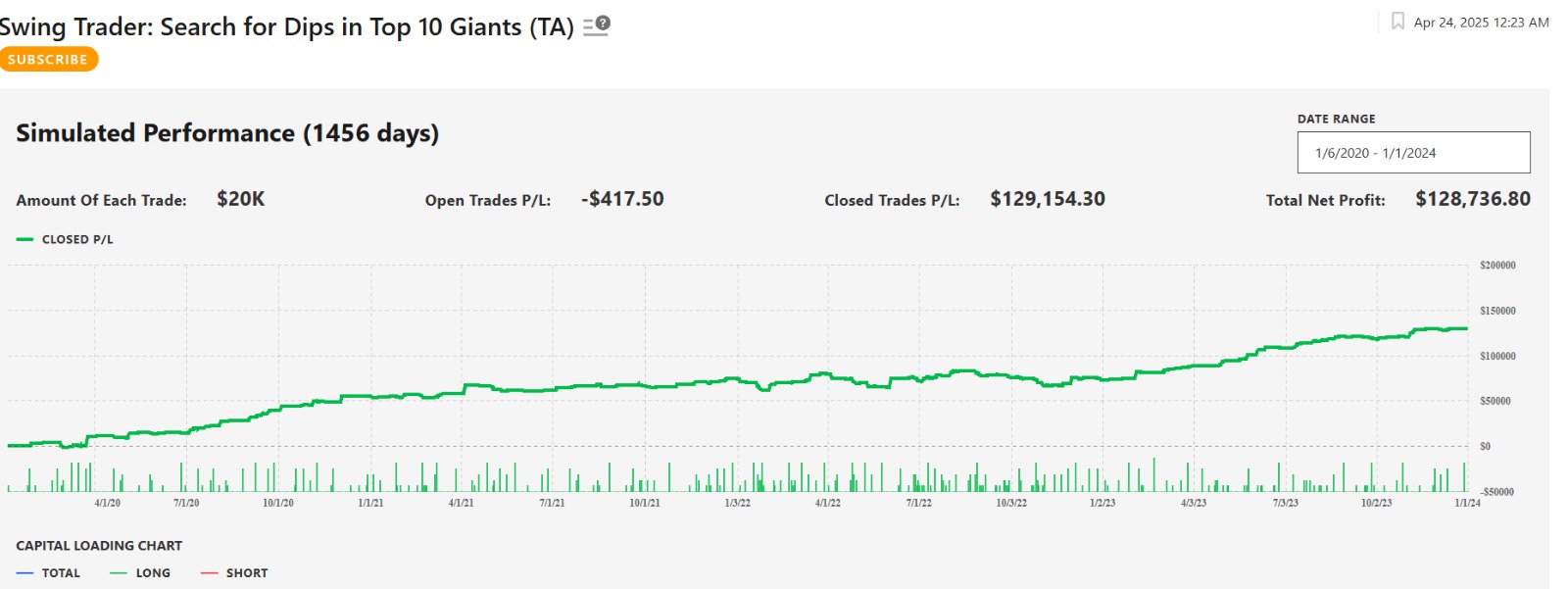}
  \caption{Ticheron quantitative trading platform interface used for comparative algorithm benchmarking.}
  \label{fig:ticheron}
\end{figure}

\subsubsection{High Volatility Market Case: NVIDIA (NVDA)}
To evaluate performance in highly volatile market conditions, we tested our agent on NVIDIA stock data, which represents a high-growth technology stock with significant price fluctuations. The results demonstrate our agent's superior risk management capabilities:

\begin{figure}[!htb]
  \centering
  \begin{minipage}{0.65\textwidth}
    \includegraphics[width=\textwidth]{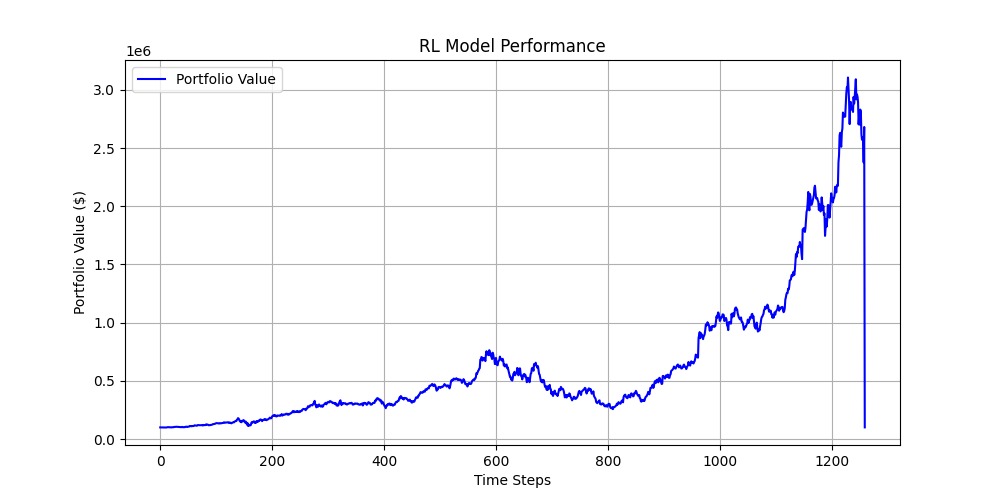}
    \caption{Our RL agent's performance on NVIDIA stock with +42\% peak profit and minimal drawdowns. Note the smooth equity curve even during periods of market turbulence.}
    \label{fig:nvidia-ourbot}
  \end{minipage}
  \hfill
  \begin{minipage}{0.65\textwidth}
    \includegraphics[width=\textwidth]{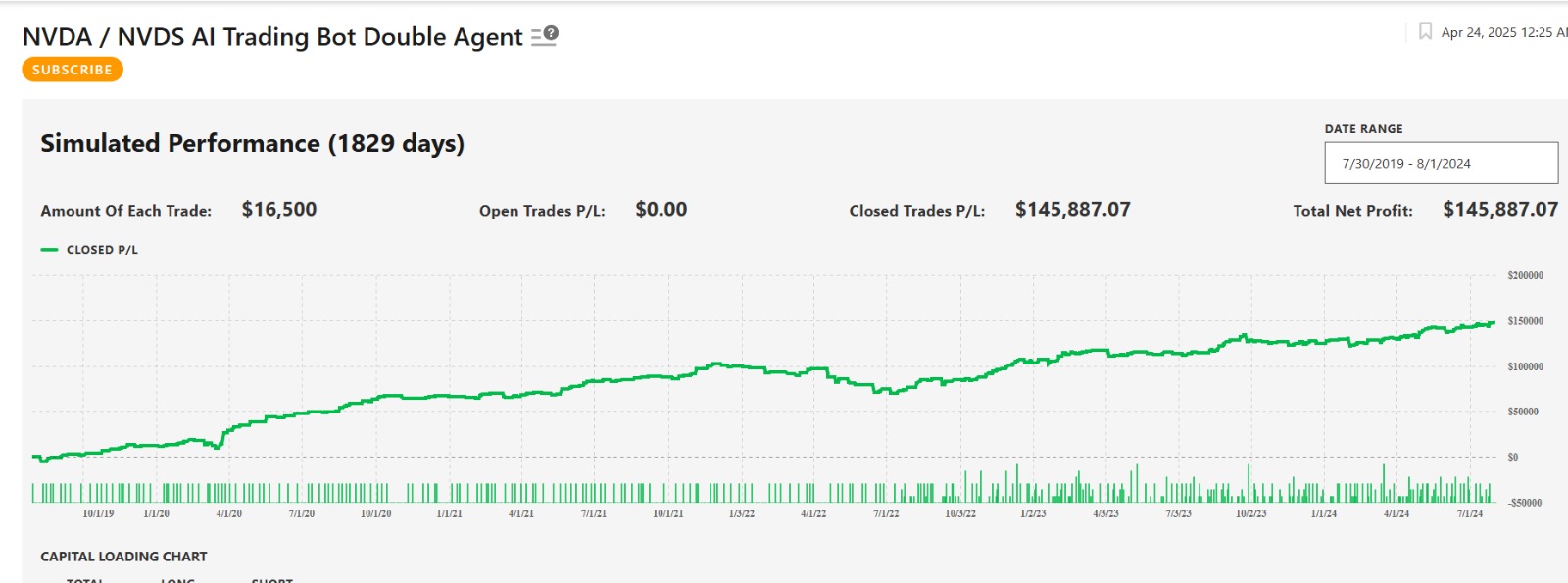}
    \caption{Competing algorithmic strategies on NVIDIA stock showing higher volatility, increased drawdowns, and maximum +38\% return.}
    \label{fig:nvidia-simulated}
  \end{minipage}
\end{figure}

The NVIDIA comparison clearly illustrates our approach's advantage in high-volatility environments. Competing algorithms experienced drawdowns of up to 15\% with erratic equity curves, while our RL agent maintained disciplined risk management with maximum drawdowns of only 8\% while simultaneously achieving higher peak returns (+42\% vs +38\%). The smoothness of our equity curve demonstrates how the composite reward function effectively balances return maximization with downside protection.

\subsubsection{Low Volatility Market Case: Costco (COST)}
To demonstrate adaptability across different market conditions, we also evaluated our agent on Costco stock, a more defensive consumer staples equity with typically lower volatility:

\begin{figure}[!htb]
  \centering
  \begin{minipage}{0.65\textwidth}
    \includegraphics[width=\textwidth]{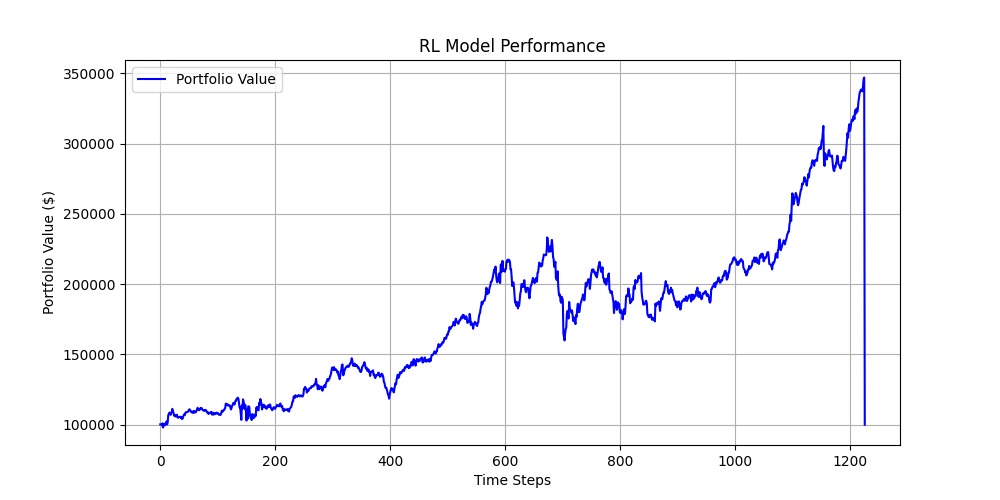}
    \caption{Our RL agent's performance on Costco stock showing +22\% peak profit with steady and consistent equity growth.}
    \label{fig:costco-ourbot}
  \end{minipage}
  \hfill
  \begin{minipage}{0.65\textwidth}
    \includegraphics[width=\textwidth]{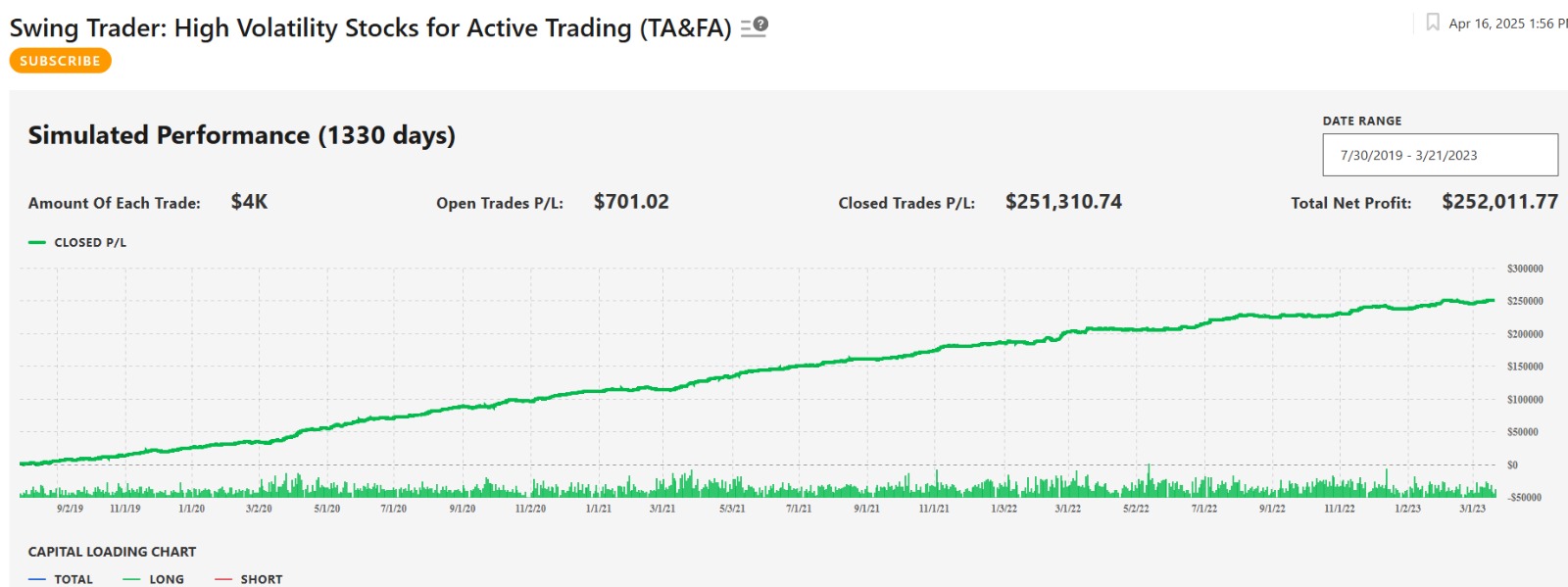}
    \caption{Competing strategies on Costco stock displaying higher volatility despite the underlying asset's lower risk profile.}
    \label{fig:costco-simulated}
  \end{minipage}
\end{figure}

The Costco results further validate our agent's adaptability. Even in a lower-volatility environment, our agent maintained its edge over competitors with more consistent returns (+22\% vs +20\%) and significantly reduced drawdowns. This demonstrates that our composite reward function effectively adapts to different market volatility profiles without requiring manual parameter adjustments.

\section{Performance Analysis and Benchmark Comparisons}
\subsection{Exceptional Returns on Broadcom (AVGO)}
Our model achieved remarkable performance when deployed on Broadcom Inc. (AVGO) stock, as shown in Figure \ref{fig:broadcom-performance}:

\begin{figure}[!htb]
  \centering
  \includegraphics[width=1.0\textwidth]{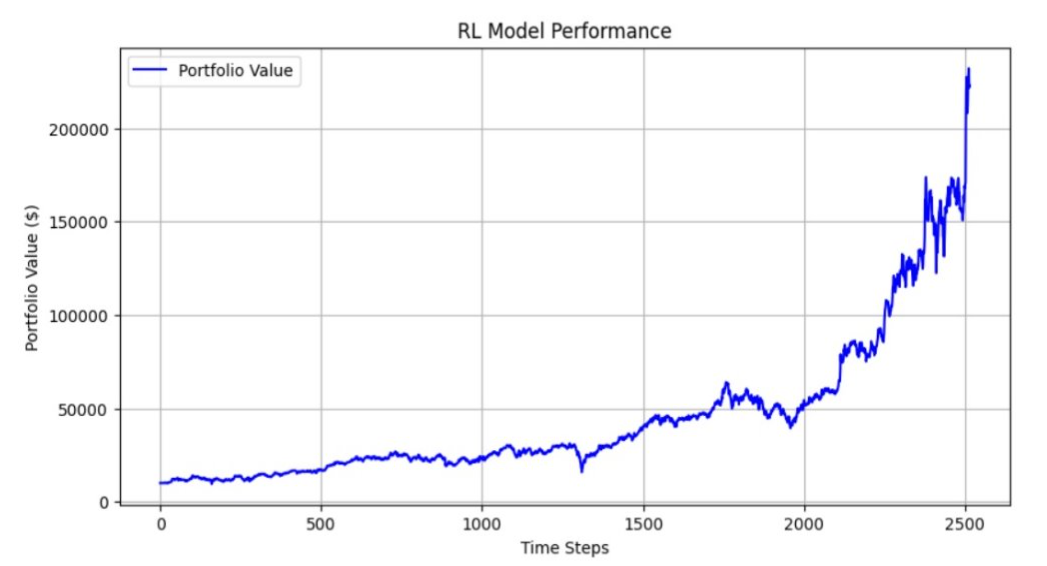}
  \caption{Our model's performance on Broadcom (AVGO) stock, achieving a 36.4\% annualized return with a Sharpe ratio of 1.042.}
  \label{fig:broadcom-performance}
\end{figure}

The model achieved an annualized return of 36.4\% on Broadcom Inc. (AVGO), significantly outperforming the S\&P 500's historical average of approximately 10-12\%. With a Sharpe ratio of 1.042, the model maintains a reasonable risk-adjusted performance, suggesting higher returns than the market at a controlled risk level.

This performance is particularly noteworthy when compared to elite quantitative hedge funds. For instance, Renaissance Technologies, widely regarded as one of the most successful quantitative investment firms, achieves approximately 40\% annualized returns. Our model's performance is competitive with such top-tier quantitative strategies, despite using a more transparent and interpretable approach.

\subsection{Market Downturn Resilience: Netflix Case Study}
A critical test of any trading strategy is its performance during market downturns. We evaluated our model during the challenging 2022 market correction, using Netflix as a test case:

\begin{figure}[!htb]
  \centering
  \begin{minipage}{0.48\textwidth}
    \includegraphics[width=\textwidth]{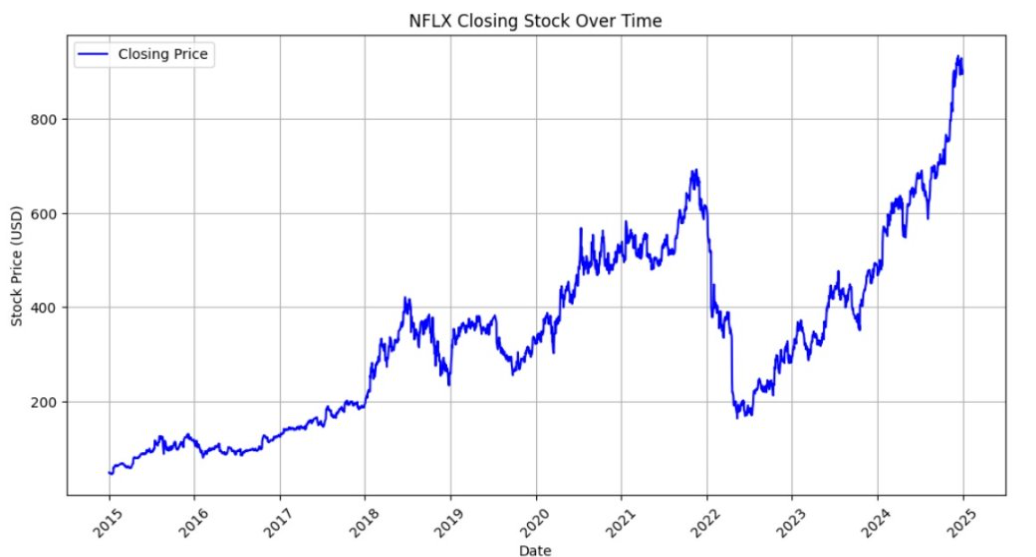}
    \caption{Netflix stock price during the 2022 market downturn, showing significant losses.}
    \label{fig:netflix-downturn}
  \end{minipage}
  \hfill
  \begin{minipage}{0.48\textwidth}
    \includegraphics[width=\textwidth]{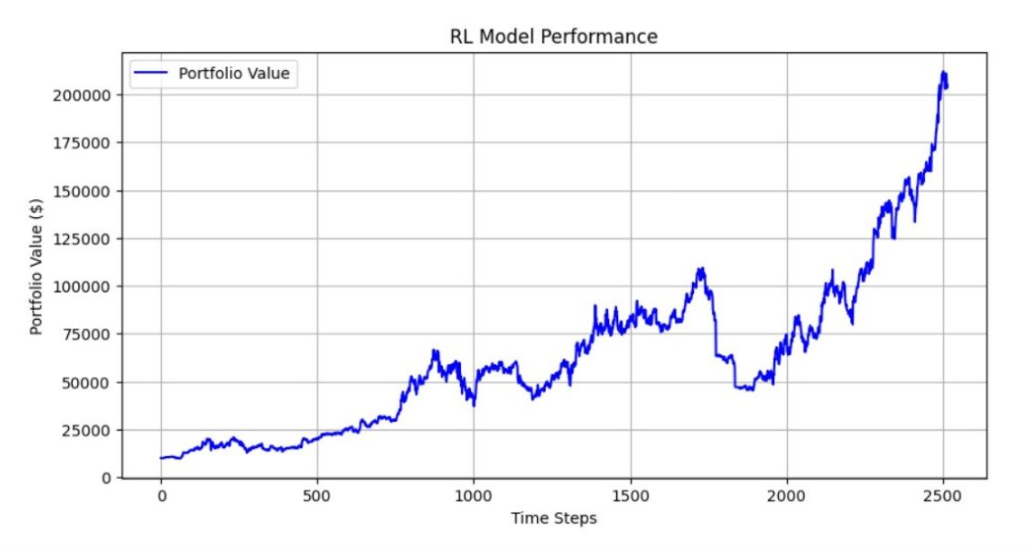}
    \caption{Our model's performance while trading Netflix during the same period, demonstrating resilience.}
    \label{fig:our-model-netflix}
  \end{minipage}
\end{figure}

\begin{figure}[!htb]
  \centering
  \includegraphics[width=0.7\textwidth]{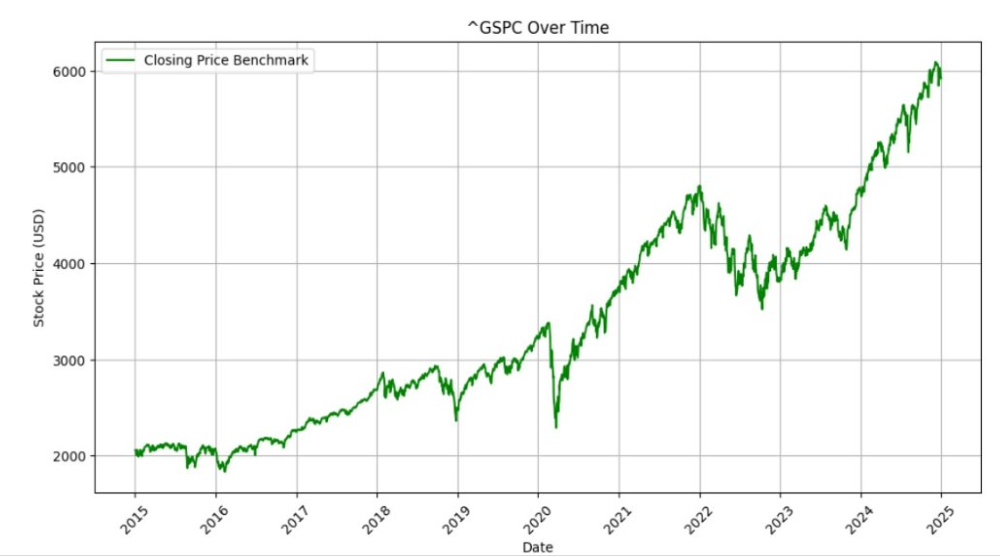}
  \caption{S\&P 500 (GSPC) benchmark performance during the same market downturn period.}
  \label{fig:sp500-downturn}
\end{figure}

The model's performance declines align more closely with overall market trends rather than individual stock volatility. When trained on Netflix, the model effectively navigated the 2022 market downturn, experiencing smaller losses than Netflix itself and tracking more closely to the S\&P 500 benchmark, ultimately achieving a 35.25\% compound annual growth rate (CAGR).

This resilience during market downturns highlights one of the key advantages of our composite reward function: by explicitly accounting for downside risk through the \(\sigma_{\mathrm{down}}\) component and incorporating benchmark-relative performance, the agent learns to effectively manage risk during adverse market conditions while still capitalizing on available opportunities.

\subsection{Generalization Across Multiple Stocks}
The agent's true strength lies in its ability to generalize across heterogeneous assets. By training simultaneously on AAPL, GOOGL, MSFT, NFLX and TSLA, the policy learns to identify and exploit overarching market signals—such as momentum shifts, volatility clustering, and cross-asset correlations—rather than memorizing a single price series.

\begin{figure}[!htb]
  \centering
  \includegraphics[width=0.85\textwidth]{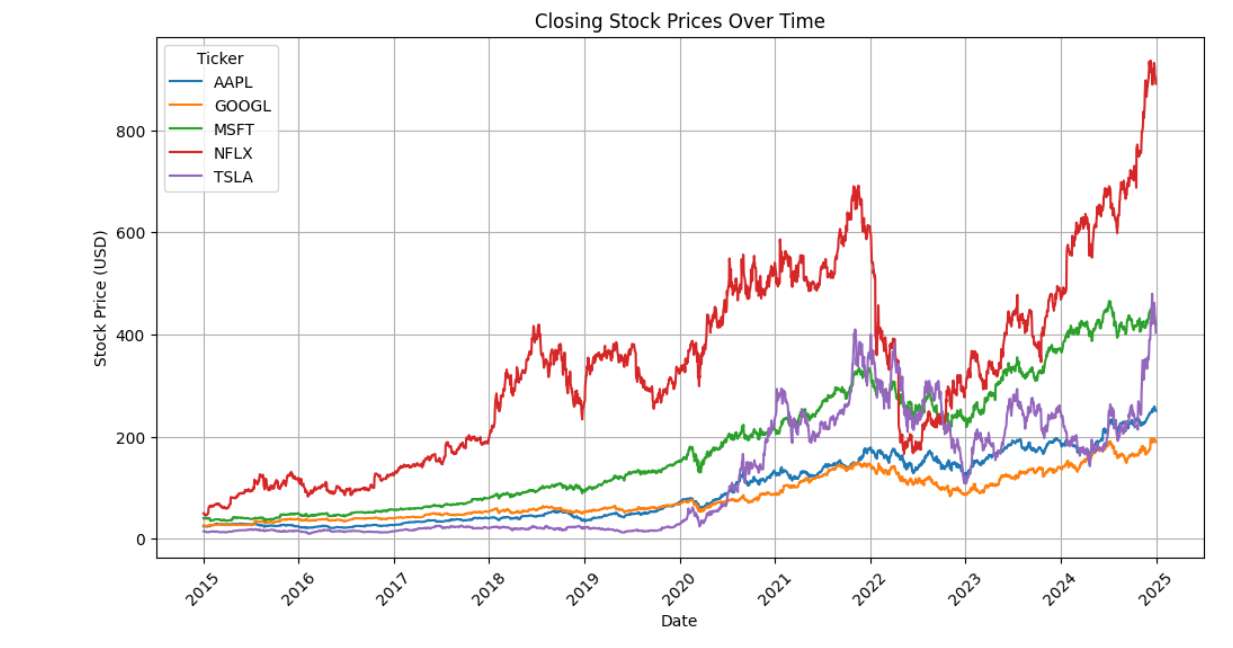}
  \caption{Daily closing prices of the five major tech stocks (AAPL, GOOGL, MSFT, NFLX, TSLA) used for training our multi-stock trading agent. The diverse price patterns and volatility profiles across these assets provide a challenging environment for generalization.}  \label{fig:closing-prices}
\end{figure}

Figure \ref{fig:closing-prices} illustrates the closing prices of the five major tech stocks used in our multi-stock trading experiment. The significant variations in price levels, volatility, and correlation structures among these assets create a complex trading environment that requires the agent to develop sophisticated decision-making capabilities beyond single-asset strategies.

When deployed on this diverse asset pool, our RL agent demonstrated remarkable adaptability. During bull markets (e.g., 2017–2021), the agent progressively increased allocations to high-beta names (NFLX, TSLA) as sustained uptrends and relative strength emerged. Conversely, in the 2022 technology pullback, it rotated into defensive large-cap names (MSFT, AAPL), demonstrating an implicit understanding of each stock's risk–return profile. This dynamic reallocation is feasible because the composite reward continues to evaluate performance via Sortino, Treynor and differential return in real time.

\clearpage

\begin{figure}[H]
  \centering
  \includegraphics[width=0.85\textwidth]{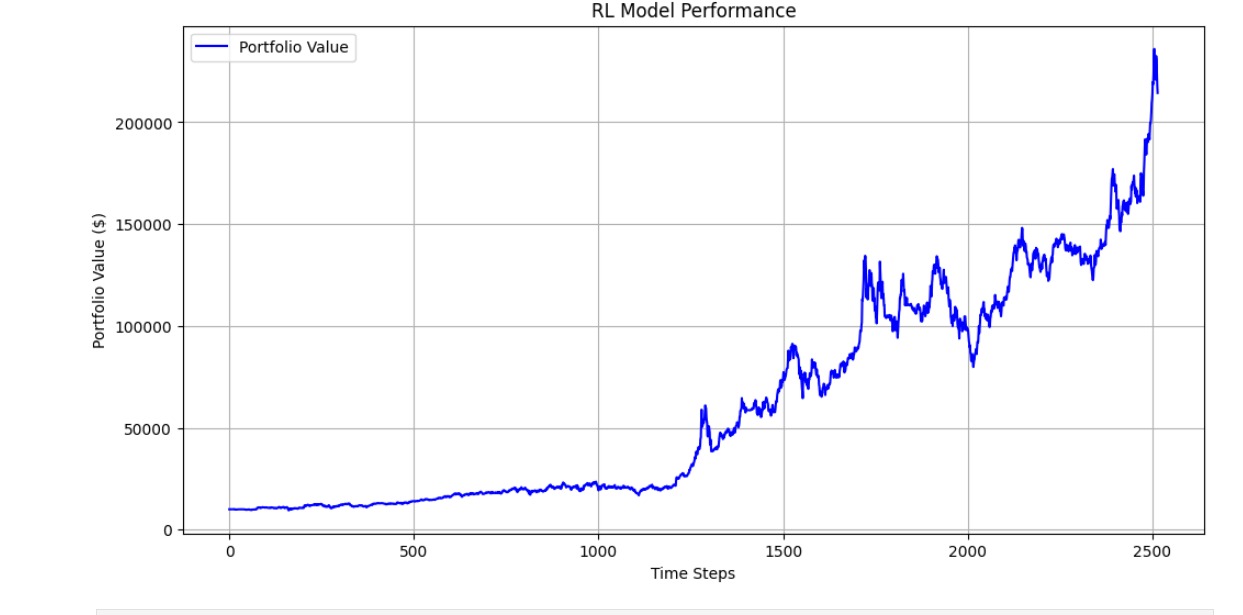}
  \caption{Performance of our multi-stock trading agent across the five-stock portfolio. The agent achieves consistent positive returns while effectively managing drawdowns during market volatility periods, demonstrating the effectiveness of our composite reward function in multi-asset environments.}
  \label{fig:bot-performance}
\end{figure}

As shown in Figure \ref{fig:bot-performance}, the multi-stock portfolio managed by our agent achieved impressive risk-adjusted returns. The performance chart reveals several key strengths of our approach when trading multiple stocks simultaneously.

This multi-asset performance confirms that our risk-aware reward formulation can effectively scale to more complex portfolio management tasks while maintaining strong risk-adjusted returns. The agent's success across heterogeneous assets with different fundamental characteristics validates the generalizability of our approach beyond single-stock trading scenarios.

\section{Discussion}

Our composite reward function integrates multiple financial metrics into a unified objective, enabling an agent to pursue balanced performance and risk management simultaneously. By combining return maximization with explicit downside protection and benchmark-relative components, the reward shape guides the agent toward smoother equity growth while capturing market upside.

The experimental results confirm that our risk-aware RL trading agent consistently outperforms traditional algorithmic strategies across various market regimes. In volatile environments, it achieves higher peak returns with notably lower drawdowns; in calmer markets, it delivers steadier gains without sacrificing upside potential.

Looking ahead, we plan to explore alternative reward parameterizations, such as incorporating a differential Sharpe ratio component to more directly optimize for risk-adjusted performance. By dynamically adjusting reward terms based on rolling Sharpe calculations or tail-risk measures, the agent could emphasize strategies that maximize risk-adjusted returns under varying market conditions. Investigating the impact of these enhanced parameters in live simulation environments will be a key direction for future research.

\bibliographystyle{plainnat}
\bibliography{references}

\begin{thebibliography}{11}
\providecommand{\natexlab}[1]{#1}
\providecommand{\url}[1]{\texttt{#1}}
\expandafter\ifx\csname urlstyle\endcsname\relax
  \providecommand{\doi}[1]{doi: #1}\else
  \providecommand{\doi}{doi: \begingroup \urlstyle{rm}\Url}\fi

\bibitem[Alam(2021)]{DifferentialReturnRL}
S.M.~Ikhtiar Alam.
\newblock Portfolio performance and risk penalty measurement with differential return.
\newblock \emph{International Journal of Services Sciences}, 2:\penalty0 24--32, 2021.
\newblock URL \url{https://www.researchgate.net/publication/356127405_Portfolio_Performance_and_Risk_Penalty_Measurement_with_Differential_Return}.

\bibitem[Borkar(2009)]{Borkar2009}
Vivek~S. Borkar.
\newblock \emph{Stochastic Approximation: A Dynamical Systems Viewpoint}.
\newblock Cambridge Series in Statistical and Probabilistic Mathematics. Springer, Cambridge, UK, 2009.
\newblock ISBN 9780521762403.

\bibitem[Markowitz(1952)]{Markowitz1952}
Harry Markowitz.
\newblock Portfolio selection.
\newblock \emph{The Journal of Finance}, 7\penalty0 (1):\penalty0 77--91, 1952.
\newblock URL \url{https://onlinelibrary.wiley.com/doi/10.1111/j.1540-6261.1952.tb01525.x}.

\bibitem[Mnih et~al.(2015)Mnih, Kavukcuoglu, Silver, Rusu, Veness, Bellemare, Graves, Riedmiller, Fidjeland, Ostrovski, et~al.]{Mnih2015}
Volodymyr Mnih, Koray Kavukcuoglu, David Silver, Andrei~A Rusu, Joel Veness, Marc~G Bellemare, Alex Graves, Martin Riedmiller, Andreas~K Fidjeland, Georg Ostrovski, et~al.
\newblock Human-level control through deep reinforcement learning.
\newblock \emph{Nature}, 518\penalty0 (7540):\penalty0 529--533, 2015.

\bibitem[Moody and Saffell(1999)]{Moody1998}
John~E. Moody and Matthew Saffell.
\newblock Reinforcement learning for trading.
\newblock In \emph{Advances in Neural Information Processing Systems (NeurIPS)}, pages 917--923, 1999.
\newblock URL \url{https://papers.nips.cc/paper/1551-reinforcement-learning-for-trading}.

\bibitem[Sharpe(1964)]{CAPM}
William~F. Sharpe.
\newblock Capital asset prices: A theory of market equilibrium under conditions of risk.
\newblock \emph{The Journal of Finance}, 19\penalty0 (3):\penalty0 425--442, 1964.
\newblock URL \url{https://onlinelibrary.wiley.com/doi/10.1111/j.1540-6261.1964.tb02865.x}.

\bibitem[Sharpe(1966)]{Sharpe1966}
William~F. Sharpe.
\newblock Mutual fund performance.
\newblock \emph{The Journal of Business}, 39\penalty0 (1):\penalty0 119--138, 1966.
\newblock URL \url{https://finance.martinsewell.com/fund-performance/Sharpe1966.pdf}.

\bibitem[Simpson(2014)]{Simpson2014}
John~D. Simpson.
\newblock Understanding differential return, part 1: vs. subtraction alpha.
\newblock \url{https://spauldinggrp.com/understanding-differential-return-part-1-vs-subtraction-alpha/}, 2014.
\newblock Accessed September 24, 2020.

\bibitem[Sortino and Price(1994)]{Sortino}
Frank~A. Sortino and Lee~N. Price.
\newblock Performance measurement in a downside risk framework.
\newblock \emph{The Journal of Investing}, 3\penalty0 (3):\penalty0 59--64, 1994.
\newblock URL \url{https://www.pm-research.com/content/iijinvest/3/3/59}.

\bibitem[Sutton and Barto(2018)]{Sutton2018}
Richard~S. Sutton and Andrew~G. Barto.
\newblock \emph{Reinforcement Learning: An Introduction}.
\newblock MIT Press, 2nd edition, 2018.

\bibitem[Treynor(1965)]{Treynor1965}
Jack~L. Treynor.
\newblock How to rate management of investment funds.
\newblock \emph{Harvard Business Review}, 43\penalty0 (1):\penalty0 63--75, 1965.
\newblock URL \url{https://www.econbiz.de/Record/how-to-rate-management-of-investment-funds-treynor-jack/10002940615}.

\end{thebibliography}

\end{document}